\documentclass[10pt,twocolumn,letterpaper]{article}

\usepackage[pagenumbers]{cvpr}
\usepackage{times}
\usepackage{epsfig}
\usepackage{graphicx}
\usepackage{amsmath}
\usepackage{amssymb}
\usepackage{array}
\usepackage{tabularray}
\usepackage{booktabs}
\usepackage{cite}
\usepackage{enumitem}
\usepackage{multirow}
\usepackage{bm}

\usepackage[pagebackref,breaklinks,colorlinks]{hyperref}
\usepackage[capitalize]{cleveref}
\crefname{section}{Sec.}{Secs.}
\Crefname{section}{Section}{Sections}
\crefname{figure}{Fig.}{Figs.}
\Crefname{figure}{Figure}{Figures}
\Crefname{table}{Table}{Tables}
\crefname{table}{Tab.}{Tabs.}

\def\iccvPaperID{11309} 

\begin{document}
\title{PointCAT:Cross-Attention Transformer for Point Cloud}

\author{Xincheng Yang\textsuperscript{1}\quad Mingze Jing\textsuperscript{1}\quad Weiji He\textsuperscript{1,}\thanks{Corresponding author}\quad Qian Chen\textsuperscript{1}\\\\
\textsuperscript{1}Jiangsu Key Laboratory of Spectral Imaging and Intelligence Sense\\
Nanjing University of Science and Technology\\
{\tt\small yaphetys@gmail.com}\quad{\tt\small \{JinMZ, hewj, chenq\}@mail.njust.edu.cn}
}
\maketitle
\begin{abstract}
  Transformer-based models have significantly advanced natural language processing and computer vision in recent years. However, due to the irregular and disordered structure of point cloud data, transformer-based models for 3D deep learning are still in their infancy compared to other methods. In this paper we present Point Cross-Attention Transformer (PointCAT), a novel end-to-end network architecture using cross-attentions mechanism for point cloud representing. Our approach combines multi-scale features via two seprate cross-attention transformer branches. To reduce the computational increase brought by multi-branch structure, we further introduce an efficient model for shape classification, which only process single class token of one branch as a query to calculate attention map with the other. Extensive experiments demonstrate that our method outperforms or achieves comparable performance to several approaches in shape classification, part segmentation and semantic segmentation tasks. The code is available at \url{https://github.com/xincheng-yang/PointCAT}
\end{abstract}

\section{Introduction}
\label{sec:intro}
3D data are becoming increasingly available and affordable with the rapid development of many areas such as autonomous driving, robotics, augmented reality and remote sensing. Unlike well structured 2D images, point clouds distributing in 3D space in an irregular and disorderd manner, which can not be processed by existing standard deep learning models directly. Multi-view based methods\cite{wei2020view,yang2019learning,su2015multi}, first project the input point clouds into multiple views and then fuse these features for accurate global representin. Many recent works\cite{maturana2015voxnet,zhou2018voxelnet} voxelize the 3D space for applying 3D discrete convolutions to point clouds. These methods either induce explicit information loss or increase massive computational and memory costs. As a pioneering work, PointNet++\cite{qi2017pointnet++} extracts pointwise features with several hierarchical shared MLPs and captures local geometric structures with max-pooling layers. A variety of follow-up approaches have been proved that such hierarchical design is able to achieve high efficiency and strong representation ability\cite{xu2018spidercnn,liu2019relation} of point cloud.
\begin{figure}
  \centering
  \includegraphics[width=\linewidth]{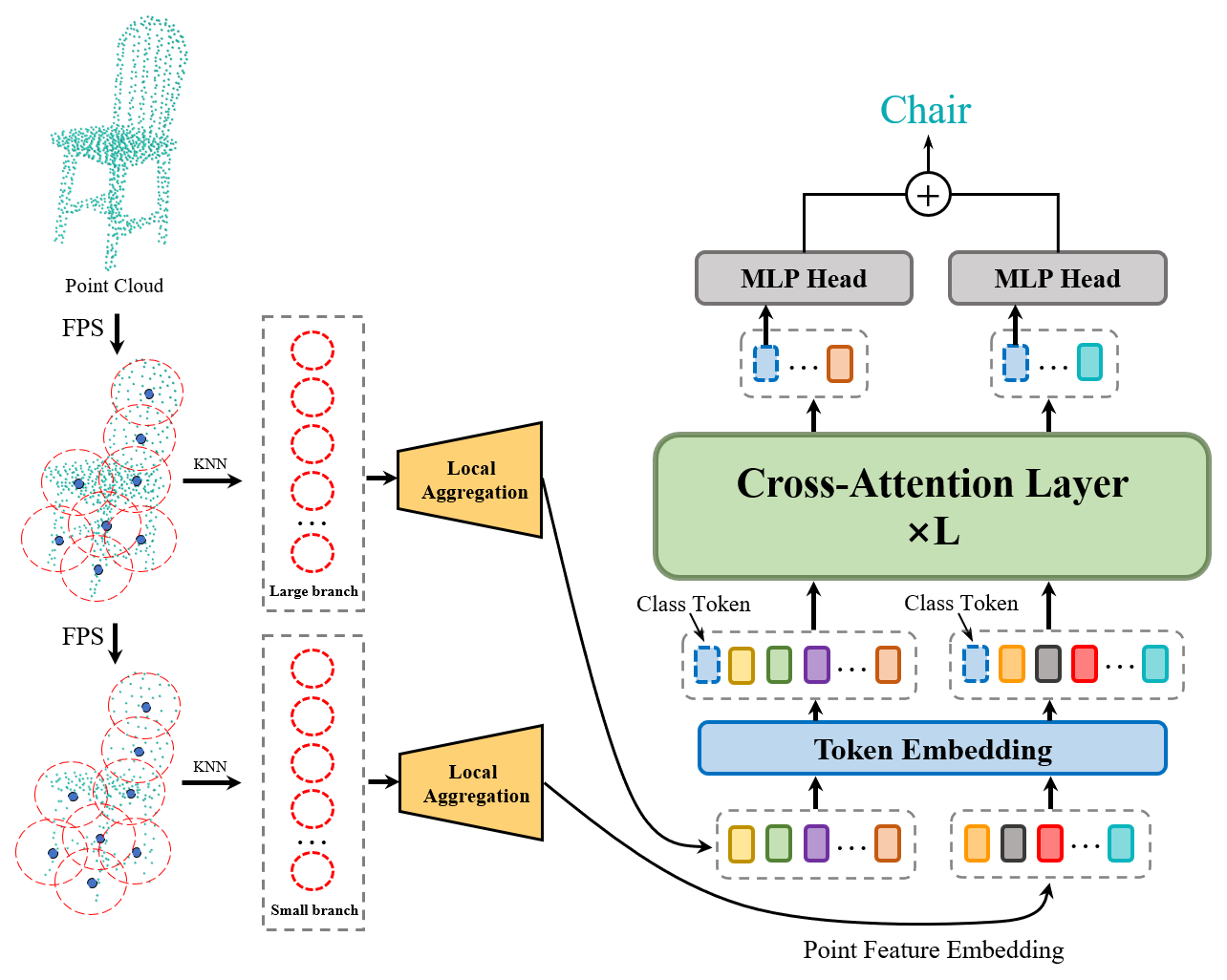}
  \caption{Illustration of our PointCAT architecture for Shape Classification. The input points are first divided into two multi-scale subset patches, local aggregations is then processed to capture high-dimensional features of point embeddings.We add an extra learnable class token to each sequence before a stack of $L$ cross-attention transformer layers. Lastly, the obtained class token is fed into MLP heads and output predictions.}
  \vspace{-0.4cm}
  \label{fig:fig1}
\end{figure}

Recently, Transformer\cite{vaswani2017attention} has gradually emerged as the dominant deep learning models and obtained great performances for most of the computer vision tasks, such as segmentation\cite{zhu2020deformable}, detection\cite{heo2021rethinking} and generation\cite{park2022styleformer}. Inspired by the signiﬁcant achievements of transformer in 2D vision tasks, substantial approaches have been developed for 3D analysis\cite{zhao2021point,guo2021pct,misra2021end}. Transformer has a natural superiority for learning semantic information of geometric features as its attention mechanism perfectly fits the sparseness, disorder and irregularity characteristics of point clouds. Based on this understanding, PointTransformer\cite{zhao2021point} verifies the feasibility of transformer model in 3D tasks for the first time. It utilizes permutation-invariant transformation by constructing Point Transformer layer for point cloud processing. PCT\cite{guo2021pct} introduces an optimized offset-attention model with implicit Laplace operators to effectively represent robust relative point cloud coordinates. Both networks leverage the order invariance of attention mechanism and made progress on several benchmarks\cite{wu20153d,chang2015shapenet,armeni20163d}.

Multi-scale feature has been proven useful for many CNN-based models\cite{cheng2020higherhrnet}. To investigate their potential benefits and construct stronger 3D transformers, we propose a novel dual-branches cross attention transformer network architecture called PointCAT, which enhances the capability of capturing long-range and multi-level dependencies among sampled points. As described in studies such as \cite{qi2017pointnet++,zhao2021point}, most common methods for feature learning involve grouping local points and aggregating local features through farthest point sampling (FPS) and k-nearest neighbor (KNN) search which only consider the local geometric information of each set but have a lack of interactions between them. To address this issue, we introduce a cross-attention layer, which processes both positional and content features through two separate transformer branches to exchange multi-scale information by attention. Based on the cross-attention layer, we construct residual Point Cross-Attention Transformer networks for several 3D understanding tasks. For a detailed introduction to the point cross-attention layer, please refer to \textbf{\Cref{sec:sec3}}. The networks are designed as hierarchical structure to avoid vast computational growth caused by cross-attention. The overall architecture of our method is illustrated in \textbf{\Cref{fig:fig1}}. Thanks to the learned long-dependency correlation, PointCAT is able to provide more precise and distinguishable features for representing point clouds. The main contributions of this paper can be summarized as follows:
\begin{itemize}[itemsep=0pt,topsep=0pt,parsep=0pt]
  \item[1)]
    We have developed an efficient hierarchical structure for extracting multi-scale feature representations in 3D understanding. Our method is able to learn precise positional information while reducing computational complexity.
    \vspace{0.1cm}
  \item[2)]
    Expanding upon such hierarchical structure, we propose a noval dual-branches cross attention transformer architecture named PointCAT, which fully combines positional and content features at different levels. It is inherently suitable for point cloud learning due to the  permutation invariance of transformer.
    \vspace{0.1cm}
  \item[3)]
    Extensive experiments demonstrate that our approach with long-range dependency enhancement performs better than or on par with several works over multiple domains and datasets.
\end{itemize}

\section{Related Work}
\label{sec:sec2}
This section briefly introduces previous deep learning networks for point cloud learning, including projection-based, voxel-based, graph-based and pointwise MLP methods. We also pay particular attention to visual attention-based works which inspire our model.
\subsection{Point Cloud Representation Learning}
\setlength{\parskip}{6pt}{
  \noindent
  \textbf{Projection-based methods.} MVCNN\cite{su2015multi} is a pioneering work of these methods. It first projects point clouds into different views of image planes, then extracts a global descriptor through mutiple 2D CNNs and max-pool layers. Yang \etal~\cite{yang2019learning} proposed a relation network to effectively exploit the inter-relationships over a group of views and integrates those views to obtain a distinguishable feature. Instead of choosing a global projection viewpoint, Tatarchenko \etal~\cite{tatarchenko2018tangent} proposed mapping local neighborhoods to local tangent planes and processing them with 2D convolutional operators. View-GCN\cite{wei2020view} uses the multiple views as graph nodes to construct a graph convolutional neural network. It gradually aggregates non-local messages over the view-graph to form the shape descriptor. FPConv\cite{lin2020fpconv} flattens each point by learned weight map to softly project surrounding points onto a regular 2D grid. While projecting an irregular point cloud to a regular 2D representation is ingenious, this process will cause the loss of 3D geometric information simultaneously. Additionally, the choice of projection methods may also heavily influence final performances.}

\setlength{\parskip}{6pt}{
  \noindent
  \textbf{Voxel-based methods.} Another technique is to voxelize the 3D space and apply 3D convolutions\cite{maturana2015voxnet,graham20183d,kuang2020voxel}. However, due to the cubic growth in the number of voxels, these strategies can bring massive computational and memory cost. To address this issue, OctNet\cite{riegler2017octnet} partitions a point cloud using hierarchical and compact octree structure which enables efficient deep networks. Qi \etal~\cite{qi2016volumetric} explored the performance differences between two types of CNNs: Voxel-based CNNs and Multi-view based CNNs, proposed a multi-resolution filter for 3D analysis. Brock \etal~\cite{brock2016generative} developed Voxception-ResNet architecture based on ResNet\cite{he2016deep}. While these methods have demonstrated good accuracy, they still lack flexbility because of the discrete properties of point clouds.}

\setlength{\parskip}{6pt}{
  \noindent
  \textbf{Graph-based methods.} Graph-based networks use a graph structure to represent point clouds. As a pioneering approach, Simonovsky \etal~\cite{simonovsky2017dynamic} treated each point of the point cloud as a vertex and created directed edges between vertices based on their neighboring points. DGCNN\cite{wang2019dynamic} used a feature space to create a graph and performed regular updates to the graph following each layer of the network. The edges were associated with feature learning functions implemented by multilayer perceptrons (MLPs). PointGCN\cite{zhang2018graph} utilized a graph that was constructed by selecting k nearest neighbors from a point cloud, with edge weights determined using a Gaussian kernel. The convolutional filters were formulated as Chebyshev polynomials in the graph spectral domain. ClusterNet\cite{chen2019clusternet} employed the unsupervised agglomerative hierarchical clustering technique to construct hierarchical structures of a point cloud and utilized a rotation-invariant module to extract rotation-invariant features for each point.}

\setlength{\parskip}{6pt}{
  \noindent
  \textbf{Pointwise MLP methods.} Researchers have proposed pointwise networks to directly process the original 3D coordinates. PointNet\cite{qi2017pointnet} is considered a milestone in 3D understanding, using several MLPs for independent pointwise feature learning and achieving permutation invariance with a max-pooling operator. In order to extract fine local geometric structures, Qi \etal~\cite{qi2017pointnet++} further introduced PointNet++, which gradually aggregates neighbor points into groups and learns high dimensional features through set abstraction layers. Based on PointNet++, Point Attention Transformer\cite{yang2019modeling} represents each point point by its absolute position and relative positions to its KNN neighbours. Yan \etal~\cite{yan2020pointasnl} utilized a robust local-nonlocal module to improve point features and proposed an Adaptive Sampling algorithm to adaptively adjust the local coordinates and relative features of sampled points. Several subsequent works have exhibited promising results\cite{qian2022pointnext,ran2021learning}. Our method follows the hierarchical structure of PointNet++ but establishes stronger and longer relation dependencies between points.}
\setlength{\parskip}{6pt}{
  \subsection{Transformers}
  The remarkable progress made by Transformer networks has attracted a lot attention in coumputer vision community\cite{dosovitskiy2020image,liu2022swin}. Transformer architectures are based on the self-attention machanism which learns weight maps between elements of sequences.}

\setlength{\parskip}{6pt}{
  \noindent
  \textbf{Vision Transformer.} Bello\etal~\cite{bello2019attention} introduced a novel self-attention mechanism to replace part of the standard convolutional operators in vision tasks. Vision Transformer(ViT)\cite{dosovitskiy2020image} splits images into patches and flattens them as the input vectors of original transformer. This was the first time that pure transformer networks showed similar performance compared to previous convolutional neural networks (CNNs). Swin Transformer\cite{liu2022swin} proposed a hierarchical Transformer with shifted windows that limits self-attention to non-overlapping local windows. CrossViT\cite{chen2021crossvit} processes dual branches of different computational complexity, largely improves the performances by 2$\%$ with only a small increase in FLOPs and model parameters.}

\setlength{\parskip}{3pt}{
  \noindent
  \textbf{Transformer for 3D Analysis.} The irregular and permutation invariant nature of point cloud makes Transformer even more suitable for 3D analysis.
  Zhao \etal~\cite{zhao2021point} investigated the application of self-attention to 3D point clouds. The Point Transformer network can serve as a general backbone for 3D understanding across domains. PCT\cite{guo2021pct} merges the positional encoding and input embeddings into a unique coordinate-based embedding module and utilizes an effective offset-attention module to capture global features. This particular optimization can be considered as a laplace process. Inspired by the masked autoencoding module, \etal~\cite{pang2022masked} generalized the concept of MAE\cite{he2022masked} to 3D point cloud and achieved some improvements.}

\section{Point Cross-Attention Transformer}
\label{sec:sec3}
In this section, we first present our method which generates dual local point feature patches through two stages: Multi-scale Grouping and Token Embedding. Then we introduce our point cross-attention transformer and explain how it integrates features of different level to provide enhanced long-dependency information. Lastly, we demonstrate the effectiveness of PointCAT architecture on various tasks of point cloud processing, including point cloud classification, part segmentation and semantic segmentation.

\subsection{Multi-scale Grouping Module}\label{sec:Multi-scale Grouping Module}
The Multi-scale Grouping Module aims to embed the input coordinates into higher-dimensional feature spaces and downsample input points to create herarchical local point groups. We first perform the farthest point sampling(FPS) and $K$-Nearest Neighbor(KNN) search to obtain sampled point groups. By aggregating these groups to their center points, the module is able to encode comprehensive regional details. Note that the diverse and sparse geometric structures in local regions may require more robust models. Therefore, we optimize the grouping process with a learnable local geometric shift.

More specifically, given input points $P=\{{{p}_{i}}|i=1,...,N\}\in {{\mathbb{R}}^{N\times 3}}$ where $N$ is the number of points with cartesian coordinates ($x$, $y$, $z$), we use linear projection for point embedding and apply farthest point sampling to choose a subset of point features ${{F}_{s}}=\{{{f}_{j}}|j=1,...,n\}\in {{\mathbb{R}}^{n\times d}}$. The local grouping process can be formulated as:
\begin{equation}
  {{F}_{g}}=\varphi (\frac{K(P,{{F}_{s}})-{{F}_{s}}}{\sigma +\varepsilon }) \label{equ:equ1}
\end{equation}
where $K$ indicates the k-nearest neighbor algorithm, $\sigma$ is the scalar standard deviation across channels, and $\varepsilon=1\times e^{-5}$ is a small float number to prevent division by zero\cite{ioffe2015batch}. $\varphi$ denotes the linear geometric shift parameters, which can be learned to improve the stability of our model. The resulting set ${{F}_{g}}=\{{{f}_{j}^{'}}|j=1,...,n\}\in {{\mathbb{R}}^{N\times k\times d}}$ represents a group of local point features that have been standardized, where $k$ is the number of neighboring points. This local normalization grouping strategy can mitigate the impact of different receiving distributions of each layer and enables deeper networks.

We follow PointNet++\cite{qi2017pointnet++}, and the local feature aggregation can be expressed as follows:
\begin{equation}
  F=\gamma (\Phi(f_{j}^{'})|j=1,...n) \label{equ:equ2}
\end{equation}
Here $\Phi$ refers to point-wise residual MLPs and $\gamma$ indicates a max-pooling operator. We double the feature channels through MLPs to construct point feature patches ${F}=\{{{x}_{i}}|i=1,...,n\}\in {{\mathbb{R}}^{N\times 2d}}$. Four blocks of this herarchical grouping structure is stacked to obtain multi-scale features of the whole point set.
\subsection{Token Embedding}
Similar to ViT's $[ class ]$ token, we prepend a learnable embedding $[{x}_{cls}]$ (which is considerd as a point cloud representation token) to the sequence of grouped point patches.
\begin{equation}
  \label{equ:equ3}
  \begin{split}
    X& =Concat([{x}_{cls}], F)  \\
    &=\{{{x}_{cls},{x}_{i}}|i=1,...,n\}\in {{\mathbb{R}}^{(N+1)\times 2d}}
  \end{split}
\end{equation}

In classification task, we feed the trained head tokens into a simple MLP for prediction. By fusing multi-scale class tokens, our method achieves both efficiency and accuracy.

\subsection{Cross-Attention Layer}
\begin{figure}[t]
  \centering
  \setlength{\abovecaptionskip}{0.3cm}
  \includegraphics[width=\linewidth]{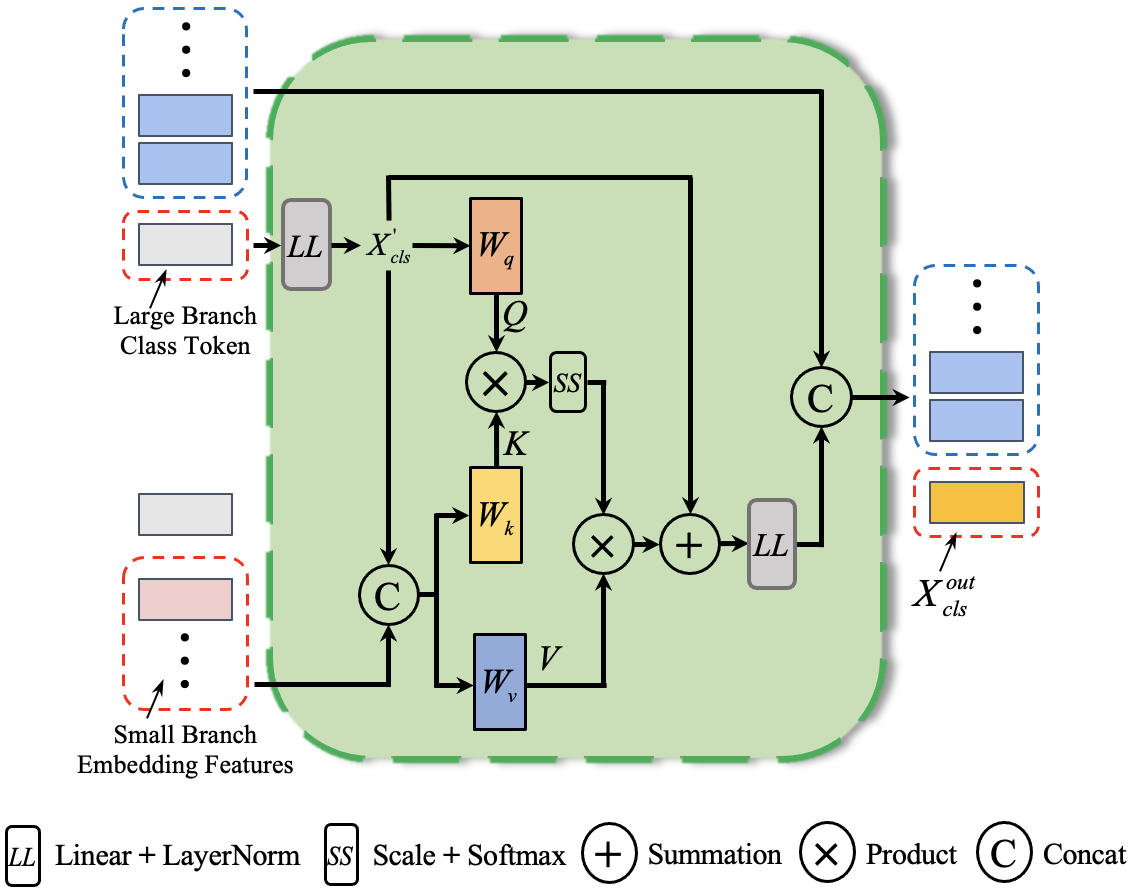}
  \caption{Cross-Attention Layer for large point branch. The ${x}_{cls}$ of large branch is projected to the feature dimension of small branch and serves as cross-attention's \textbf{Query} to interact with the \textbf{Key} and \textbf{Value} generated from the embedding feature tokens of the small branch. We align the dimensions back to the large branch through another linear projection with Layernorm. Ultimately, we concatenate the processed class token with original patch tokens.}
  \label{fig:fig2}
  \vspace{0.1cm}
\end{figure}

Our Cross-Attention Layer is illustrated in \textbf{\Cref{fig:fig2}}. As shown in the figure, given the input dual-branch point patches $X_{L}^{l-1}=\{{X}_{cls}^{l-1}, X_{patch}^{l-1}\}$ and $Y_{S}^{l-1}=\{{Y}_{cls}^{l-1}, Y_{patch}^{l-1}\}$, we obtain new point features of the $l$-th large branch by following operations:
\begin{equation}
  \begin{aligned}
     & {X}_{cls}^{l'}=LN(Linear({X}_{cls}^{l-1}))             \\
     & X^{l}=Concat\{{X}_{cls}^{l'}, {Y}_{patch}^{l-1}\}      \\
     & {X}_{cls}^{l}=MSA(X^{l}) + {X}_{cls}^{l'}              \\
     & {X}_{L}^{l}=Concat\{{X}_{cls}^{l}, {X}_{patch}^{l-1}\}
  \end{aligned}
\end{equation}
each layer contains a multihead self-attention (MSA) layer. $Linear$ represents linear projection to align point feature dimensions. Layer normalization (LN) and residual shortcuts are applied during feature dimension fusion. The small branch follows the same procedure but processes the class token $Y_{cls}^{l-1}$ and patch tokens $X_{patch}^{l-1}$ instead. The cross-attention mechanism can be expressed as:
\begin{equation}
  \begin{aligned}
     & Q =\bm{W_q} \cdot {X}_{cls}^{l'},K=\bm{W_k} \cdot X^l,V=\bm{W_v} \cdot X^l \\
     & MSA(Q,K,V) = {\rm{Softmax}}(\frac{{Q{K^T}}}{{\sqrt{{d_k}}}})V
  \end{aligned}
\end{equation}
where ${W_q, W_k, W_v\in {{\mathbb{R}}^{c \times {d_k}}}}$ are learnable weight matrixes, $c$ and ${d_k}$ here represent the feature channels and dimensions of each head. The class token ${X}_{cls}^{l-1}$ is able to fully interact with patch features from the other branch through this cross-attention design. Compared to the original Transformer, we do not apply any feed-forward networks. In addition, there is no need for position encoding because point clouds already contain the location information itself.

We stacked these cross-attention layers to further boost the performance of our network. As a result, PointCAT is able to learn discriminative global descriptors for point clouds with geometrically and long-dependency information, making it a powerful backbone for various point cloud tasks.


\subsection{Implementation Details}
This section elucidates how we implemented PointCAT and highlights the small differences between the architectures used for shape classification, part segmentation and semantic segmentation tasks.

\setlength{\parskip}{2pt}{
  \noindent
  \textbf{Shape Classification.} For shape classification, the architecture only needs to predict a global feature vector for all points. As mentioned in \textbf{\Cref{sec:Multi-scale Grouping Module}}, each stage of Multi-Scale Grouping Module reduces the number of points by half and doubles the feature dimensions. We feed the output point features of the last two stages into a Cross-Attention Transformer Layer to obtain multi-scale global representations. Then, two separate MLPs classify the multi-scale feature tokens respectively, and lastly we we sum the results for the final overall classification prediction.}

\setlength{\parskip}{2pt}{
  \noindent
  \textbf{Part Segmentation.} We follow the propagation strategy of PointNet++\cite{qi2017pointnet++} to gradually propagate features from subsampled points to the original set of points. Unlike the classification architecture, the global feature is aggregated by max-pooling operation. In addition, we encode the one-hot category label as a 64-dimensional vector and concatenate it with the global feature. Finally, we concatenate the class token, propagted points and global feature as a whole, applying an MLP to map this feature to the final logits.}

\setlength{\parskip}{2pt}{
  \noindent
  \textbf{Semantic Segmentation.} Semantic segmentation architecture is basically the same as the part segmentation architecture, with the biggest difference being that the semantic segmentation network fully utilizes color information of points in S3DIS\cite{armeni20163d} as an additional modality characteristic of the point clouds.}
\section{Experiments}
\label{sec:sec4}
In this section, we comprehensively evalute the performance of PointCAT on three public datasets: ModelNet40\cite{wu20153d} for shape classification, ShapeNetPart\cite{chang2015shapenet} for part Segmentation and S3DIS\cite{armeni20163d} for semantic segmentation. To provide better clarity of our PointCAT, we also report ablation studies that were conducted on ModelNet40 and S3DIS dataset.
\begin{table}[ht]
  \center
  \resizebox{\columnwidth}{!}{%
    \begin{tabular}{@{}l|cccc@{}}
      \toprule
      Method                                     & Inputs               & mAcc(\%)      & OA(\%) & Vote       \\ \midrule
      \textbf{Other Methods}                     & \multicolumn{4}{c}{}                                       \\ \midrule
      PointNet\cite{qi2017pointnet}              & 1k points            & 86.0          & 89.2   & ×          \\
      PointNet++\cite{qi2017pointnet++}          & 1k points            & --            & 90.7   & ×          \\
      PointNet++\cite{qi2017pointnet++}          & 5k points+normal     & --            & 91.9   & ×          \\
      SO-Net\cite{li2018so}                      & 2k points+normal     & --            & 90.9   & ×          \\
      PointCNN\cite{li2018pointcnn}              & 1k points            & 88.1          & 92.5   & \checkmark \\
      DGCNN\cite{wang2019dynamic}                & 1k points            & 90.2          & 92.9   & ×          \\
      DensePoint\cite{liu2019densepoint}         & 1k points            & --            & 93.2   & \checkmark \\
      RSCNN\cite{liu2019relation}                & 1k points            & --            & 92.9   & \checkmark \\
      KD-Net\cite{klokov2017escape}              & 32k points           & --            & 91.8   & ×          \\
      KPConv\cite{thomas2019kpconv}              & 1k points            & --            & 92.9   & ×          \\
      Point2Sequence\cite{liu2019point2sequence} & 1k points            & 90.4          & 92.6   & ×          \\
      \midrule
      \textbf{Transformer-based Methods}         & \multicolumn{4}{c}{}                                       \\ \midrule
      3DETR\cite{misra2021end}                   & 2k points+normal     & 89.9          & 91.9   & ×          \\
      GAPNeT\cite{chen2021gapointnet}            & 1k points            & 89.7          & 92.4   & ×          \\
      PointANSL\cite{yan2020pointasnl}           & 1k points            & --            & 92.9   & ×          \\
      PointTransformer\cite{zhao2021point}       & 1k points            & 90.6          & 93.7   & \checkmark \\
      PCT\cite{guo2021pct}                       & 1k points            & --            & 93.2   & ×          \\ \midrule
      PointCAT                                   & \textbf{1k points}   & \textbf{90.9} & 93.5   & \textbf{×} \\ \bottomrule
    \end{tabular}%
  }
  \caption{Classification results on ModelNet40 dataset.}
  \label{tab:table1}
  \vspace{-0.4cm}
\end{table}

\subsection{Shape Classification}
\noindent
\textbf{Dataset and Metrics.} ModelNet40\cite{wu20153d} consists of 12311 synthetic models with 40 object categories. We follow the widly used data division that splits ModelNet40 into 9843 instances for training and 2468 for testing. All models were randomly downsampled to 1024 points. During training, A random anisotropic scaling in [0.8, 1.2], a translation in range [-0.2, 0.2] and a random input dropout are applied for data augmentation. During testing, we do not use any extra augmentation or voting methods. For evaluation metrics, we utilize the mean accuracy operated on each category (mAcc) and the overall accuracy (OA) operated on all classes.

\noindent
\textbf{Performance comparison.} Experimental results are shown in \textbf{\Cref{tab:table1}}. The overall accuracy on ModelNet40 is 93.5\%, which outperforms almost all state-of-the-art methods in 1k input points expect PointTransformer\cite{zhao2021point} which uses a tricky voting strategy. In addition, PointCAT achieves Top-1 $mAcc$(90.9\%) which means our model is more robust compared to other approaches. Note that the current model does not consider any point normals as inputs, so we may further improve the performance theoretically.

\subsection{Part Segmentation}
\setlength{\parskip}{2pt}{
  \noindent
  \textbf{Dataset and Metrics.} We also evaluate our methods on ShapeNetPart\cite{chang2015shapenet} for part segmentation. ShapeNetPart contains 16881 objects covering 16 categories and 50 part labels, each instance has 2 to 6 parts. Following PointNet++\cite{qi2017pointnet++}, we randomly selecte 2048 points as the inputs for each object. The data augmentations are set the same as the training of shape classification task, but without the inputs dropout. The evaluation metrics we used include three componets: category mIoU, instance mIoU and the IoU for each category.}

\setlength{\parskip}{2pt}{
  \noindent
  \textbf{Performance comparison.} \textbf{\Cref{tab:table2}} lists the part segmentation results on ShapeNetPart. We compare our methods with several recent works, including DGCNN\cite{wang2019dynamic}, SPLATNet\cite{su2018splatnet}, etc. The result show that our approach achieves a competitive performance on both class mIoU and instance mIoU. Moreover, PointCAT outperforms all prior models as measured by the IoU for some categories such as cap, aerphone, rocket and table.}

\begin{figure}[ht]
  \centering
  \includegraphics[width=\columnwidth]{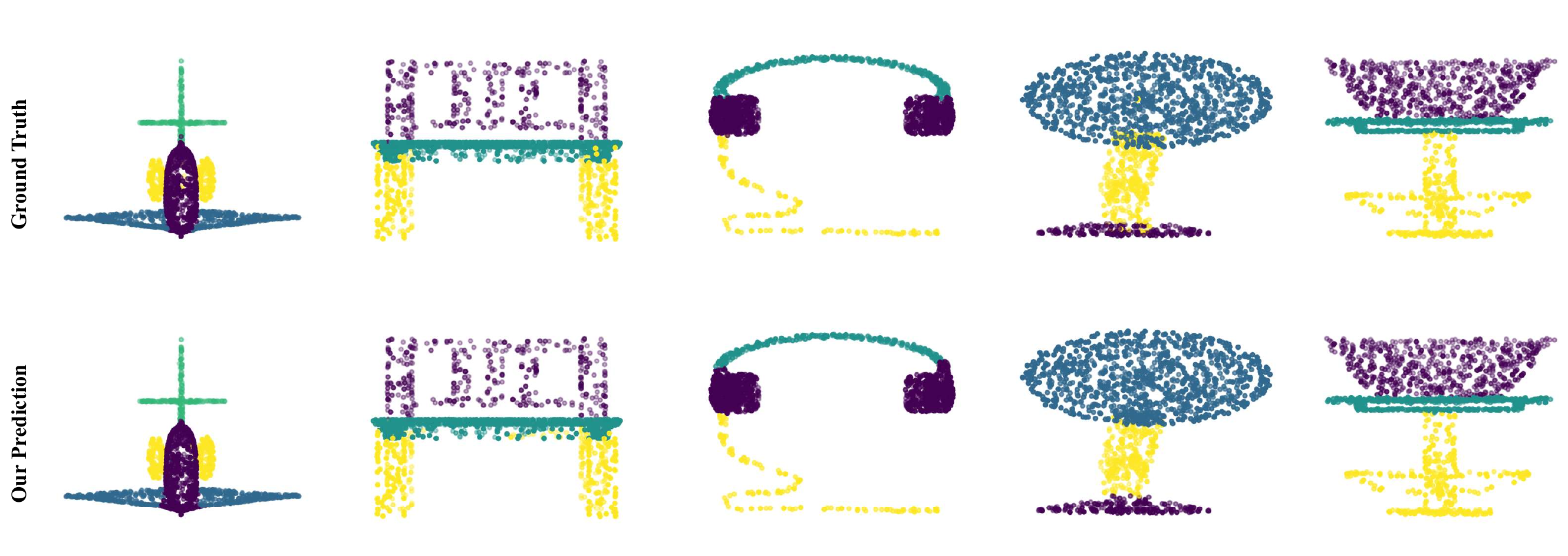}
  \caption{Visualization results on ShapeNetPart. Top line shows the ground truth and bottom line is our prediction.}
  \label{fig:fig3}
  \vspace{-0.2cm}
\end{figure}

\setlength{\parskip}{2pt}{
  \noindent
  \textbf{Visualization.} Object part segmentation results are visualized in \textbf{\Cref{fig:fig3}}. As shown in the picture, we represent different parts of the same object in different colors. The predictions of our PointCAT are close to the ground truth, which intuitively exhibits the strong point cloud understanding ability of our methods.}

\begin{table*}[ht]
  \centering
  \resizebox{\textwidth}{!}{%
    \begin{tabular}{l|cc|cccccccccccccccc}
      \hline
      \multicolumn{1}{c|}{Method}            & \begin{tabular}[c]{@{}c@{}}Cls.\\ mIoU\end{tabular} & \begin{tabular}[c]{@{}c@{}}Inst.\\ mIoU\end{tabular} & aero          & bag           & cap           & car           & chair         & \begin{tabular}[c]{@{}c@{}}aerp-\\ hone\end{tabular} & guitar        & knife         & lamp          & laptop        & \begin{tabular}[c]{@{}c@{}}motor-\\ bike\end{tabular} & mug           & pistol        & rocket        & \begin{tabular}[c]{@{}c@{}}skate-\\ board\end{tabular} & table         \\ \hline
      PointNet\cite{qi2017pointnet}          & 80.4                                                & 83.7                                                 & 83.4          & 78.7          & 82.5          & 74.9          & 89.6          & 73.0                                                 & 91.5          & 85.9          & 80.8          & 95.3          & 65.2                                                  & 93.0          & 81.2          & 57.9          & 72.8                                                   & 80.6          \\
      PointNet++\cite{qi2017pointnet++}      & 81.9                                                & 85.1                                                 & 82.4          & 79.0          & 87.7          & 77.3          & 90.8          & 71.8                                                 & 91.0          & 85.9          & 83.7          & 95.3          & 71.6                                                  & 94.1          & 81.3          & 58.7          & 76.4                                                   & 82.6          \\
      P2Sequence\cite{liu2019point2sequence} & --                                                  & 85.2                                                 & 82.6          & 81.8          & 87.5          & 77.3          & 90.8          & 77.1                                                 & 91.1          & 86.9          & 83.9          & 95.7          & 70.8                                                  & 94.6          & 79.3          & 58.1          & 75.2                                                   & 82.8          \\
      PointCNN\cite{li2018pointcnn}          & 84.6                                                & 86.1                                                 & 84.1          & \textbf{86.5} & 86.0          & 80.8          & 90.6          & 79.7                                                 & \textbf{92.3} & 88.4          & 85.3          & 96.1          & 77.2                                                  & 95.2          & 84.2          & 64.2          & 80.0                                                   & 83.0          \\
      RS-CNN\cite{liu2019relation}           & 84.0                                                & 86.2                                                 & 83.5          & 84.8          & 88.8          & 79.6          & 91.2          & 81.1                                                 & 91.6          & \textbf{88.4} & \textbf{86.0} & 96.0          & 73.7                                                  & 94.1          & 83.4          & 60.5          & 77.7                                                   & 83.6          \\
      PCNN\cite{wang2018deep}                & 81.8                                                & 85.1                                                 & 82.4          & 80.1          & 85.5          & 79.5          & 90.8          & 73.2                                                 & 91.3          & 86.0          & 85.0          & 95.7          & 73.2                                                  & 94.8          & 83.3          & 51.0          & 75.0                                                   & 81.8          \\
      SpiderCNN\cite{xu2018spidercnn}        & 82.4                                                & 85.3                                                 & 83.5          & 81.0          & 87.2          & 77.5          & 90.7          & 76.8                                                 & 91.1          & 87.3          & 83.3          & 95.8          & 70.2                                                  & 93.5          & 82.7          & 59.7          & 75.8                                                   & 82.8          \\
      DGCNN\cite{wang2019dynamic}            & 82.3                                                & 85.2                                                 & 84.0          & 83.4          & 86.7          & 77.8          & 90.6          & 74.7                                                 & 91.2          & 87.5          & 82.8          & 95.7          & 66.3                                                  & 94.9          & 81.1          & 63.5          & 74.5                                                   & 82.6          \\
      SPLATNet\cite{su2018splatnet}          & 83.7                                                & 85.4                                                 & 83.2          & 84.3          & 89.1          & 80.3          & 90.7          & 75.5                                                 & 92.1          & 87.1          & 83.9          & \textbf{96.3} & 75.6                                                  & \textbf{95.8} & 83.8          & 64.0          & 75.5                                                   & 81.8          \\
      PointMLP\cite{ma2022rethinking}        & \textbf{84.6}                                       & 86.1                                                 & 83.5          & 83.4          & 87.5          & 80.5          & 90.3          & 78.2                                                 & 92.2          & 88.1          & 82.6          & 96.2          & \textbf{77.5}                                         & 95.8          & \textbf{85.4} & 64.6          & \textbf{83.3}                                          & 84.3          \\
      PointASNL\cite{yan2020pointasnl}       & --                                                  & 86.1                                                 & 84.1          & 84.7          & 87.9          & 79.7          & \textbf{92.2} & 73.7                                                 & 91.0          & 87.2          & 84.2          & 95.8          & 74.4                                                  & 95.2          & 81.0          & 63.0          & 76.3                                                   & 83.2          \\
      SO-Net\cite{li2018so}                  & --                                                  & 84.9                                                 & 82.8          & 77.8          & 88.0          & 77.3          & 90.6          & 73.5                                                 & 90.7          & 83.9          & 82.8          & 94.8          & 69.1                                                  & 94.2          & 80.9          & 53.1          & 72.9                                                   & 83.0          \\
      Kd-Net\cite{klokov2017escape}          & --                                                  & 82.3                                                 & 80.1          & 74.6          & 74.3          & 70.3          & 88.6          & 73.5                                                 & 90.2          & 87.2          & 81.0          & 94.9          & 57.4                                                  & 86.7          & 78.1          & 51.8          & 69.9                                                   & 80.3          \\
      PCT\cite{guo2021pct}                   & --                                                  & \textbf{86.4}                                        & \textbf{85.0} & 82.4          & 89.0          & \textbf{81.2} & 91.9          & 71.5                                                 & 91.3          & 88.1          & 86.3          & 95.8          & 64.6                                                  & 95.8          & 83.6          & 62.2          & 77.6                                                   & 83.7          \\ \hline
      PointCAT                               & 84.4                                                & 86.0                                                 & 83.0          & 83.8          & \textbf{90.1} & 79.8          & 90.2          & \textbf{83.4}                                        & 91.8          & 87.8          & 82.5          & 95.9          & 76.1                                                  & 95.4          & 84.9          & \textbf{68.5} & 83.1                                                   & \textbf{84.1} \\ \hline
    \end{tabular}%
  }
  \setlength{\belowcaptionskip}{0.2cm}
  \caption{Part segmentation results on ShapeNetPart dataset. We report the mean IoU across all part classes Cls.mIoU(\%), the mean IoU across all instances Inst.mIoU(\%) and the IoU(\%) for each category. Compareed to the results reported in the cited papers, our method achieves a competitive performance.}
  \label{tab:table2}
  \vspace{-0.3cm}
\end{table*}

\subsection{Semantic Segmentation}
\vspace{-0.1cm}
\noindent
\textbf{Dataset and Metrics.} Large-scale real-world scene segmentation is a more challenging task. We further assess our method on S3DIS(Stanford Large-Scale 3D Indoor Spaces Dataset), which covers six large-scale indoor areas from three different buildings for a total of 273 million points annotated with 13 classes(ceiling, floor, table, etc). For a fair comparison, the same data processing method as PointNet++\cite{qi2017pointnet++} is utilized. We used the most difficult Area 5 for testing, and it was not included in the training process. For evaluation metrics, we calculate the mean classwise intersection over union (mIoU), mean of classwise accuracy (mAcc), and overall pointwise accuracy (OA).

\noindent
\textbf{Performance comparison.} We evaluated our approach on the S3DIS's hardests scene area 5, the results are shown in \textbf{\Cref{tab:table4}}. Our method achieves a competitive value of mean accuracy (mAcc) score. However, in terms of overall accuracy (OA) and mean intersection over union (mIoU), PointCAT performs even better, achieving 87.5\%\verb|\|63.8\%, which surpasses all methods. For example, we outperform MLPs-based networks such as PointNet++\cite{qi2017pointnet++}, CNN-based architectures such as PointCNN\cite{li2018pointcnn}, graph-based networks such as PointWeb\cite{zhao2019pointweb}, previous transformer-based methods such as PCT\cite{guo2021pct}. Notably, all the methods we compared against followed the same dataset partition as PointNet++\cite{qi2017pointnet++}.

\begin{figure*}[!htbp]
  \centering
  \includegraphics[width=\textwidth]{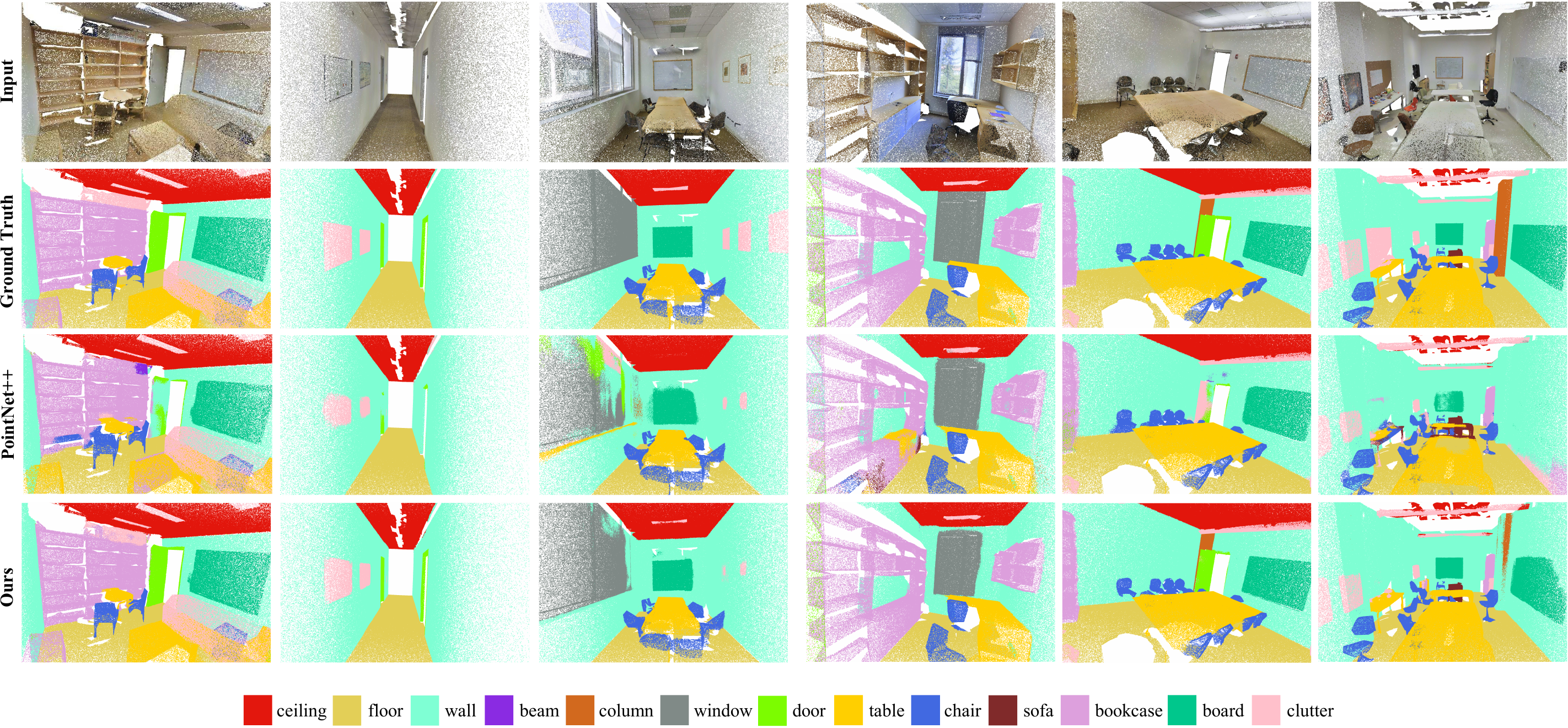}
  \vspace{0.05cm}
  \caption{Visualization of semantic segmentation results on the S3DIS dataset.}
  \label{fig:fig4}
\end{figure*}

\noindent
\textbf{Visualization.} The visualization of semantic segmentation results is shown in \textbf{\Cref{fig:fig4}}. We present the input point cloud and ground truth in the first two lines. The last two lines depict the predictions from PointNet++ (the best model provided in their original paper) and our PointCAT. It is clear that our predictions are distinguishable and close to the ground truth intuitively. Our approach achieves the best IoU in many categories, including ceiling, bookcase, board, sofa, etc.

\subsection{Ablation Study}
\vspace{0.2cm}
We conducted a series of ablation studies to verify the effectiveness of the main components in PointCAT. In order to draw more convincing conclusions, we made evaluations on both ModelNet40\cite{wu20153d} and S3DIS\cite{armeni20163d} datasets.

\begin{table}[!htbp]
  \centering
  \resizebox{0.85\columnwidth}{!}{%
    \begin{tabular}{@{}cc|ccc@{}}
      \toprule
      \multicolumn{2}{c|}{\textbf{\begin{tabular}[c]{@{}c@{}}Multi-scale\\ Grouping\end{tabular}}} & mAcc(\%) & OA(\%) & $\Delta$(\%)        \\ \midrule
      \multirow{3}{*}{$d=4$}                                                                       & $k=8$    & 86.4   & 90.3         & -3.2 \\
                                                                                                   & $k=16$   & 87.9   & 91.9         & -1.6 \\
                                                                                                   & $k=32$   & 88.4   & 92.0         & -1.5 \\    \midrule
      \multirow{3}{*}{$d=2$}                                                                       & $k=8$    & 85.8   & 90.8         & -2.7 \\
                                                                                                   & $k=16$   & 88.0   & 92.6         & -0.9 \\
                                                                                                   & $k=32$   & 90.9   & 93.5         & --   \\    \bottomrule
    \end{tabular}%
  }
  \caption{Ablation study of multi-scale grouping strategy.}
  \label{tab:table3}
  \vspace{0.2cm}
\end{table}

\begin{table*}[!htbp]
  \centering
  \resizebox{\textwidth}{!}{%
    \begin{tabular}{@{}l|ccc|ccccccccccccc@{}}
      \toprule      Method               & OA            & mAcc & mIoU          & ceiling       & floor & wall & beam & column & window & door & table         & chair & sofa          & bookcase      & board         & clutter       \\ \midrule
      PointNet\cite{qi2017pointnet}      & --            & 49.0 & 41.1          & 88.8          & 97.3  & 69.8 & 0.1  & 3.9    & 46.3   & 10.8 & 59.0          & 52.6  & 5.9           & 40.3          & 26.4          & 33.2          \\
      PointNet++\cite{qi2017pointnet++}  & 83.4          & 64.2 & 54.3          & 90.6          & 97.0  & 75.6 & 0.0  & 6.3    & 58.1   & 19.4 & 69.4          & 74.8  & 51.5          & 62.2          & 58.7          & 42.9          \\
      SEGCloud\cite{tchapmi2017segcloud} & --            & 57.4 & 48.9          & 90.1          & 96.1  & 69.9 & 0.0  & 18.4   & 38.4   & 23.1 & 70.4          & 72.9  & 40.9          & 58.4          & 13.0          & 41.6          \\
      PointCNN\cite{li2018pointcnn}      & 85.9          & 63.9 & 57.3          & 92.3          & 98.2  & 79.4 & 0.0  & 17.6   & 22.8   & 62.1 & 74.4          & 80.6  & 31.7          & 66.7          & 62.1          & 56.7          \\
      PCNN\cite{atzmon2018point}         & --            & 67.0 & 58.3          & 92.3          & 96.2  & 75.9 & 0.3  & 6.0    & 69.5   & 63.5 & 66.9          & 65.6  & 47.3          & 68.9          & 59.1          & 46.2          \\
      HPEIN\cite{jiang2019hierarchical}  & 87.2          & 68.3 & 61.9          & 91.5          & 98.2  & 81.4 & 0.0  & 23.3   & 65.3   & 40.0 & 75.5          & 87.7  & 58.5          & 67.8          & 65.6          & 49.4          \\
      PAT\cite{yang2019modeling}         & --            & 70.8 & 60.1          & 93.0          & 98.5  & 72.3 & 1.0  & 41.5   & 85.1   & 38.2 & 57.7          & 83.6  & 48.1          & 67.0          & 61.3          & 33.6          \\
      DGCNN\cite{wang2019dynamic}        & --            & 84.1 & 56.1          & --            & --    & --   & --   & --     & --     & --   & --            & --    & --            & --            & --            & --            \\
      PointWeb\cite{zhao2019pointweb}    & 87.0          & 66.6 & 60.3          & 92.0          & 98.5  & 79.4 & 0.0  & 21.1   & 59.7   & 34.8 & 76.3          & 88.3  & 46.9          & 69.3          & 64.9          & 52.5          \\
      PCT\cite{guo2021pct}               & --            & 67.7 & 61.3          & 92.5          & 98.4  & 80.6 & 0.0  & 19.3   & 61.6   & 48.0 & 76.6          & 85.2  & 46.2          & 67.7          & 67.9          & 52.3          \\ \midrule
      PointCAT                           & \textbf{88.2} & 71.0 & \textbf{64.0} & \textbf{94.2} & 98.3  & 80.5 & 0.0  & 18.6   & 55.5   & 58.9 & \textbf{77.2} & 88.0  & \textbf{64.8} & \textbf{72.2} & \textbf{68.9} & \textbf{55.4} \\ \bottomrule
    \end{tabular}%
  }
  \setlength{\abovecaptionskip}{0.2cm}
  \caption{Semantic segmentation results, tested on Area 5 of the S3DIS dataset. All results of other methods are taken from cited papers.}
  \label{tab:table4}
  \vspace{0.2cm}
\end{table*}

\begin{table}[!htbp]
  \centering
  \resizebox{\columnwidth}{!}{%
    \begin{tabular}{@{}cc|cccccc@{}}
      \toprule
      \multicolumn{2}{c|}{\multirow{2}{*}{\textbf{\begin{tabular}[c]{@{}c@{}}Cross-Attention\\ Layer\end{tabular}}}} &
      \multirow{2}{*}{mAcc}                                                                                          &
      \multirow{2}{*}{OA}                                                                                            &
      \multirow{2}{*}{mIoU}                                                                                          &
      \multirow{2}{*}{FLOPs}                                                                                         &
      \multirow{2}{*}{Params}                                                                                        &
      \multirow{2}{*}{TE}                                                                                                                                             \\
      \multicolumn{2}{c|}{}                                                                                          &     &      &      &      &      &              \\ \midrule
      \multicolumn{1}{c|}{\multirow{2}{*}{\textbf{S3DIS}}}                                                           & MSA & 67.2 & 85.9 & 61.9 & 7.4  & 87.0 & 78.0  \\
      \multicolumn{1}{c|}{}                                                                                          & CA  & 71.0 & 88.2 & 64.0 & 3.7  & 54.4 & 70.0  \\ \midrule
      \multicolumn{1}{c|}{\multirow{2}{*}{\textbf{ModelNet40}}}                                                      & MSA & 88.7 & 92.0 & --   & 10.6 & 87.6 & 113.0 \\
      \multicolumn{1}{c|}{}                                                                                          & CA  & 90.9 & 93.5 & --   & 8.9  & 33.1 & 102.0 \\ \bottomrule
    \end{tabular}%
  }
  \setlength{\abovecaptionskip}{0.2cm}
  \caption{Ablation study of cross-attention layer. We report mAcc (\%), OA(\%), mIoU(\%), FLOPs(G), Params(M) and Time per Epoch(s) on both S3DIS and ModelNet40.}
  \label{tab:table5}
  \vspace{-0.2cm}
\end{table}

\noindent
\textbf{Multi-scale Grouping.} The first step in our method is to find a proper grouping strategy. We compared two diffrernt downsampling ratios $d$ with three different settings of $k$-neighbors on the ModelNet40\cite{wu20153d} dataset. The contrasting results are presented in \textbf{\Cref{tab:table3}}, where $\Delta$ indicates the variation of $OA$. When $d$ was set to 4, there was a notable drop in both $OA$ and $mAcc$. The best performance was achieved when we set $d$ to 2 and $k$ to 32 respectively, which allowed us to obtain two branches with 128 and 64 points by repeating the grouping process four times. However, the performance kept getting worse as $k$ decreased. Under these circumstances of relatively small $k$, only a limited number of data points were available for each cross-attention layer so that PointCAT did not have enough contextual information to make accurate predictions.

\noindent
\textbf{Cross-Attention Layer.} To demonstrate the effectiveness of the dual-branch cross-attention layer design, we conducted experiments where the point patches were processed only with multi-head self-attention transformers on ModelNet40 and S3DIS. To ensure a fair comparison, we kept other network components and parameters the same. As presented in \textbf{\Cref{tab:table5}}, the performance gap between the standard transformer and our cross-attention transformer is signiﬁcant. The latter reduced the error from 8.0\% to 6.5\% on classification task and improves the mAcc/OA/mIoU(\%) from 67.2/85.9/61.9 to 71.0/88.2/64.0 on segmentation task, indicating the superiority of the cross-attention mechanism. In terms of efficiency, on S3DIS, our cross-attention layer led to a reduction of 50\% FLOPs and 37\% parameters, accelerating the inference procedure by 8 seconds per epoch. Furthermore, thanks to the class token design, PointCAT significantly reduced the model parameters from 87.6M to 33.1M when performing the classification task.

\begin{figure*}[!htbp]
  \centering
  \includegraphics[width=\textwidth]{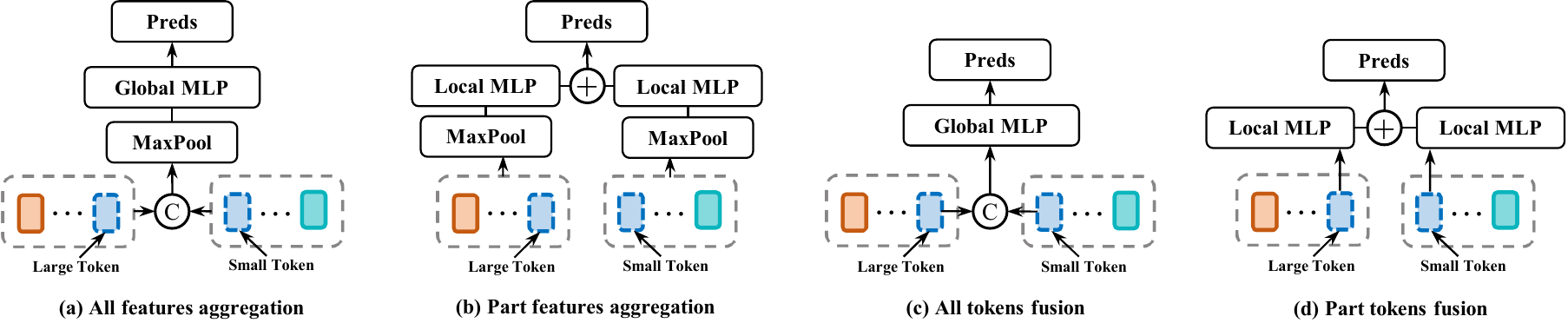}
  \caption{Multi-scale feature fusion. (a) All features aggregation. We concatenate all the features from both branches and aggregate them into a global descriptor using a max-pooling layer. The final predictions are obtained through a global classifier. (b) Part features aggregation. We handle local features from each branch separately and add both local predictions as the output. (c) All tokens fusion. We consider the class tokens from both branches as one global feature. (d) Part tokens fusion. Class tokens taken from dual branches are regarded as two global features.}
  \label{fig:fig5}
  \vspace{-0.2cm}
\end{figure*}

\begin{table}[!htbp]
  \centering
  \begin{tabular}{@{}cl|cccl@{}}
    \toprule
    \multicolumn{2}{c|}{\multirow{2}{*}{\textbf{Fusion methods}}} &
    \multirow{2}{*}{mAcc}                                         &
    \multirow{2}{*}{OA}                                           &
    \multirow{2}{*}{FLOPs}                                        &
    \multicolumn{1}{c}{\multirow{2}{*}{Params}}                                                              \\
    \multicolumn{2}{c|}{}                                         &      &      &     & \multicolumn{1}{c}{} \\ \midrule
    \multicolumn{2}{c|}{All features}                             &
    \multicolumn{1}{c}{90.5}                                      &
    \multicolumn{1}{c}{92.5}                                      &
    \multicolumn{1}{c}{11.4}                                      & 58.7
    \\
    \multicolumn{2}{c|}{Part features}                            & 90.2 & 92.9 & 8.8 & 46.2                 \\
    \multicolumn{2}{c|}{All tokens}                               & 89.9 & 92.7 & 8.6 & 30.7                 \\
    \multicolumn{2}{c|}{Part tokens}                              & 90.9 & 93.5 & 8.9 & 33.1                 \\ \bottomrule
  \end{tabular}%
  \caption{Ablation studies of fusion methods. We report mAcc (\%), OA(\%), FLOPs(G) and Params(M).}
  \label{tab:table6}
  \vspace{-0.6cm}
\end{table}

\noindent
\textbf{Feature fusion methods.} \textbf{\Cref{fig:fig5}} illustrates different fusion approaches. We compare the performances on ModelNet40 and show the results in \textbf{\Cref{tab:table6}}. Among all the compared schemes, our proposed part tokens fusion achieves the best mean and overall accuracy with a small increase in computational costs. It is interesting to note that all features aggregation fails to achieve better performance with extra information, which means class tokens are more refine and distinguishable compared to all features. Moreover, we observed that considering each branch as an independent model and simply aggregating their predictions leads to superior results.

\section{Conclusion}
In this paper, we present a noval dual-branch transformer architecture termed as PointCAT. Our approach begins by leveraging hierarchical residual MLPs to extract multi-scale groups of local points, then we utilize cross-attention layers for point cloud feature learning as they are permutation-invariant. By integrating both positional and contextual information into advanced semantic features, we significantly enhance the ability of the model to capture long-range point cloud relationships. Comprehensive experimental results on various benchmarks illustrate the outstanding point cloud representing capability of our approach.

Transformer-based methods have made tremendous strides in advancing 3D comprehension and yielded promising outcomes. We hope that our research can serve as a catalyst for the development of transformer-based networks. In future work on PointCAT, we will continue to broaden its applications to more complex domains, including point cloud generation and 3D object detection.
\vspace{-0.2cm}
{\small
  \bibliographystyle{ieee_fullname}
  \bibliography{Vertax}
}

\end{document}